\documentclass[10pt,twocolumn,letterpaper]{article}

\usepackage{cvpr}              

\usepackage[dvipsnames]{xcolor}

\definecolor{cvprblue}{rgb}{0.21,0.49,0.74}
\usepackage{multirow}
\usepackage[pagebackref,breaklinks,colorlinks,citecolor=cvprblue,]{hyperref}
\usepackage[accsupp]{axessibility} 
\def\paperID{10265} 
\def\confName{CVPR}
\def\confYear{2024}

\title{Behind the Veil: Enhanced Indoor 3D Scene Reconstruction with Occluded Surfaces Completion}
\author{
Su Sun\thanks{Equally contributed as co-first author. This work was done during Su's internship at Bosch.}~~$^{1}$, Cheng Zhao\footnotemark[1]~~$^{2}$, Yuliang Guo$^{2}$, Ruoyu Wang$^{2}$, Xinyu Huang$^{2}$, Yingjie Victor Chen$^{1}$, Liu Ren$^{2}$ \\
$^{1}$Purdue University, $^{2}$Bosch Research North America, Bosch Center for Artificial Intelligence (BCAI)\\
{\tt\small \{sun931, victorchen\}@purdue.edu }\\
{\tt\small \{cheng.zhao, yuliang.guo2, ruoyu.wang, xinyu.huang, liu.ren\}@us.bosch.com}
}

\begin{document}
\maketitle

\begin{abstract}
    \vspace{-4mm}
    In this paper, we present a novel indoor 3D reconstruction method with occluded surface completion, given a sequence of depth readings.     
    Prior state-of-the-art~(SOTA) methods only focus on the reconstruction of the visible areas in a scene, neglecting the invisible areas due to the occlusions, e.g., the contact surface between furniture, occluded wall and floor.
    Our method tackles the task of completing the occluded scene surfaces, resulting in a complete 3D scene mesh.
    The core idea of our method is learning 3D geometry prior from various complete scenes to infer the occluded geometry of an unseen scene from solely depth measurements. 
    We design a coarse-fine hierarchical octree representation coupled with a dual-decoder architecture, i.e., Geo-decoder and 3D Inpainter, which jointly reconstructs the complete 3D scene geometry.
    The Geo-decoder with detailed representation at fine levels is optimized online for each scene to reconstruct visible surfaces. 
    The 3D Inpainter with abstract representation at coarse levels is trained offline using various scenes to complete occluded surfaces.
    As a result, while the Geo-decoder is specialized for an individual scene, the 3D Inpainter can be generally applied across different scenes.
    We evaluate the proposed method on the 3D Completed Room Scene (3D-CRS) and iTHOR datasets, significantly outperforming the SOTA methods by a gain of 16.8\% and 24.2\% in terms of the completeness of 3D reconstruction.
    3D-CRS dataset including a complete 3D mesh of each scene is provided on project webpage\footnote{\url{https://github.com/BoschRHI3NA/3D-CRS-dataset}}.
    \vspace{-6mm}
\end{abstract}

    



\section{Introduction}
\label{sec:intro}
\begin{figure}[!t]
    \centering
    \includegraphics[width=0.45\textwidth]{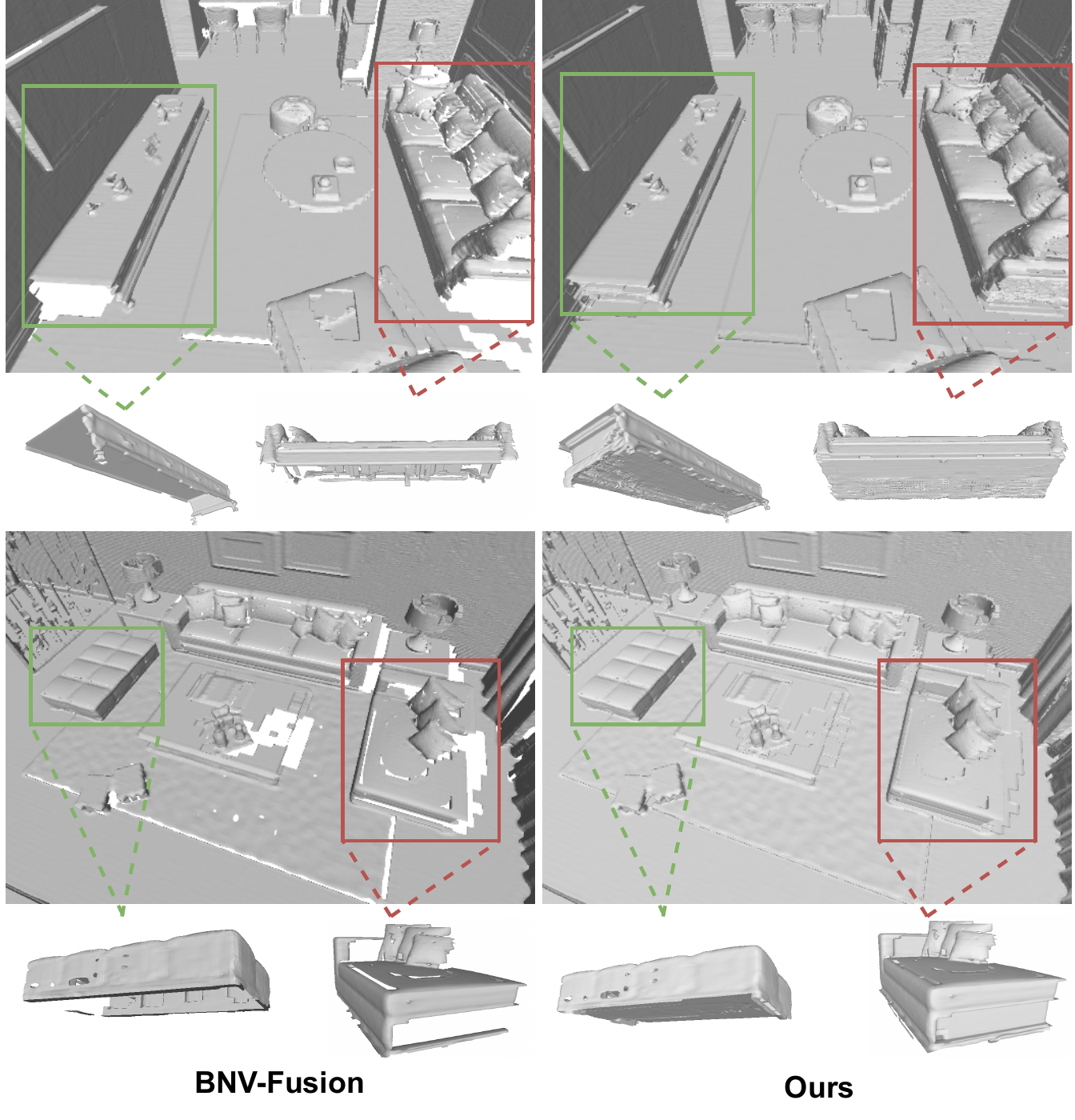}
    \vspace{-4mm}
    \captionof{figure}{\textbf{Occluded Surface Completion:} 
    Our approach marks a novel 3D surface reconstruction, uniquely completing occluded surfaces in areas invisible to existing methods. It enables more accurate 3D modelling of furniture and the reconstruction of furniture-obscured room regions, significantly advancing indoor scene reconstruction.}
    \label{fig:main:overview}
    \vspace{-8mm}
\end{figure}
%
%
%

Interactable 3D scene reconstruction at room scale is an enabling technology for a large variety of AR/VR and embodied AI applications, e.g., virtual touring, room re-arrangement, teleoperation etc. 
Over recent decades, numerous methods~\cite{curless1996volumetric, pfister2000surfels, azinovic2022neural,li2022bnv, dai2017bundlefusion, whelan2015elasticfusion} have been developed to reconstruct 3D surfaces. 
These methods have now become highly automated and capable of producing high-quality outputs. 
However, a critical aspect in advancing scene representation—making it editable, re-configurable, and suitable for object manipulation in mixed reality and embodied AI applications—lies in completing occluded areas. 
In mixed reality scenarios, for instance, it is crucial to avoid the appearance of occluded parts like the backside of a sofa or the holes on the floors during the repositioning of furniture in a living room. 
Ensuring a complete 3D shape of the furniture and a seamless layout is essential for effective object manipulation and scene editing.
However, most existing methods~\cite{azinovic2022neural,li2022bnv, dai2017bundlefusion, whelan2015elasticfusion} fall short in addressing this issue, as they predominantly focus on reconstructing visible regions where sensor measurements are available. 
This paper specifically addresses reconstructing visible surfaces and completing invisible portions in a unified framework.

The effectiveness of classic surface reconstruction methods such as TSDF~\cite{curless1996volumetric} and Surfels~\cite{pfister2000surfels} suffer from the inherent limitations of consumer-level depth sensors, such as depth noise and limited range capabilities.
These restrictions often lead to incomplete geometries such as holes in 3D reconstructions, due to the lack of depth measurements caused by sensor noise and blocked views.
In contrast, the rise of neural implicit representation methods has shown promise in addressing these challenges. 
Neural Radiance Fields (NeRF)~\cite{mildenhall2021nerf} advances neural implicit representations for shape modeling~\cite{park2019deepsdf, peng2020convolutional} and 3D reconstruction~\cite{azinovic2022neural,li2022bnv}. 
Since their internal representation is a continuous field, it naturally improves the handling of the measurement noise and incompleteness to some extent.
However, these implicit-style methods~\cite{azinovic2022neural,li2022bnv} primarily focus on reconstructing visible surfaces, often overlooking invisible surfaces obscured by occlusions, such as the furniture backside and the contact areas between furniture and floor. 
The large occluded areas are beyond the implicit representation's extrapolation capabilities without specific designs.

Recent studies have combined large 2D generative models with NeRF techniques to create complete objects~\cite{Melas-Kyriazi_2023_CVPR} or scenes~\cite{Kim_2023_CVPR} based on RGB image inputs. 
However, these focus either on object-level completion~\cite{Melas-Kyriazi_2023_CVPR} or on generating outdoor scenes without editability~\cite{Kim_2023_CVPR}, sometimes at the compromise of reconstruction quality. 
However, they are limited in accurately recovering 3D scene geometries even with RGB-based large foundation models. 
Our method prioritizes reconstructing high-quality scene geometry with a small model trained on a modest amount of data.

Our framework aims to reconstruct clean and complete 3D surfaces in room-scale scenes using only depth readings, as shown in Figure~\ref{fig:main:overview}.
The success of our method lies upon a hierarchical octree representation of a scene, coupled with a special design of dual-decoder architecture. 
Compared to existing methods~\cite{takikawa2021nglod,li2022bnv}, we uniquely treat visible and occluded surfaces from two different aspects:
1) We separately represent visible and occluded regions by fine-level and coarse-level features:
Fine features are expected to encode high-frequency detailed geometry of visible surfaces, while coarse features are anticipated to represent contextual structure information, which is more generalizable for occluded surface completion.
2) We design a scene-specific visible decoder, and a generalizable cross-scene occlusion decoder:
The visible surface decoder is optimized online using depth readings in a testing scene, whereas the occluded surface decoder is trained offline with multiple scenes.

Our main contributions can be summarized as follows: 
\begin{itemize}
\item We introduce a novel 3D surface reconstruction framework to realize occluded surface completion at scene-level;
\item We design a coarse-fine feature representation mechanism coupled with a dual-decoder architecture, which respectively reconstructs the occluded and visible geometry in a scene;
\item We design a cross-scene trained 3D Inpainter, and experimentally demonstrate its generalization ability on the occluded surface completion in unseen scenes;
%
\end{itemize}

\section{Related Work}
In the fields of computer vision and graphics, scene representations can be categorized into two primary categories: explicit and implicit representations. 
Classic explicit geometric representations~\cite{endres20133, whelan2015elasticfusion, newcombe2011kinectfusion, dai2017bundlefusion}, encompassing point clouds, voxels, and triangular meshes, have been extensively employed due to their inherent simplicity and adaptability. 
More recently, there has been growing interest in using deep implicit representations for representing 3D shapes and scenes. With this approach, 3D surface geometries are encoded into an implicit function by neural networks. 
NeRF~\cite{mildenhall2021nerf} was among the first instances to represent a scene using neural implicit representations. 
However, NeRF was initially designed for novel view synthesis, and its performance in 3D reconstruction is limited. 
The following research~\cite{azinovic2022neural, li2022bnv, Yu2022MonoSDF} leveraged implicit representation to address the incompleteness of surface reconstruction caused by the range sensor limitation, such as noise depth readings.
Neural RGB-D~\cite{azinovic2022neural} surface reconstruction demonstrates that integrating depth measurements into the radiance field formulation can yield more detailed and complete 3D surface reconstructions. 
BNV-Fusion~\cite{li2022bnv} fuses local-level neural volumes into a global neural volume, enhancing both global completeness and fine-grained reconstruction quality.
MonoSDF~\cite{Yu2022MonoSDF} demonstrates significant improvements in surface reconstruction by utilizing a pre-trained depth estimation model, which provides complementary reconstruction cues in addition to monocular images.
Despite these advancements, a common limitation persists: while these methods mitigate issues in reconstructing small missing regions, they struggle with large occluded surface completion.

Alternatively, some research has introduced object-centric representations~\cite{wu2023objectsdf++, bokhovkin2023neural, Li_2023_ICCV} or local region representations~\cite{wang2023lp, ye2023self} to indirectly enhance scene surface completion.
However, the scalability of these methods is limited due to their reliance on category-specific or region-specific priors, which often require pre-training with specific 3D model datasets. 
This requirement poses a significant challenge in covering all categories or local regions that could be present in a general indoor scene.
In addition, iMap~\cite{sucar2021imap} and NICE-SLAM~\cite{zhu2022nice} improve dense 3D reconstruction by combining the advantages of neural implicit representation with the geometry representation of the 3D SLAM system.  
However, all these methods exhibit a limited capacity for completing the surface of occluded areas at 3D scene level.

\section{Methodology}
\vspace{-1mm}
%
\begin{figure}[!t]
    \centering
    \includegraphics[width=0.4\textwidth]{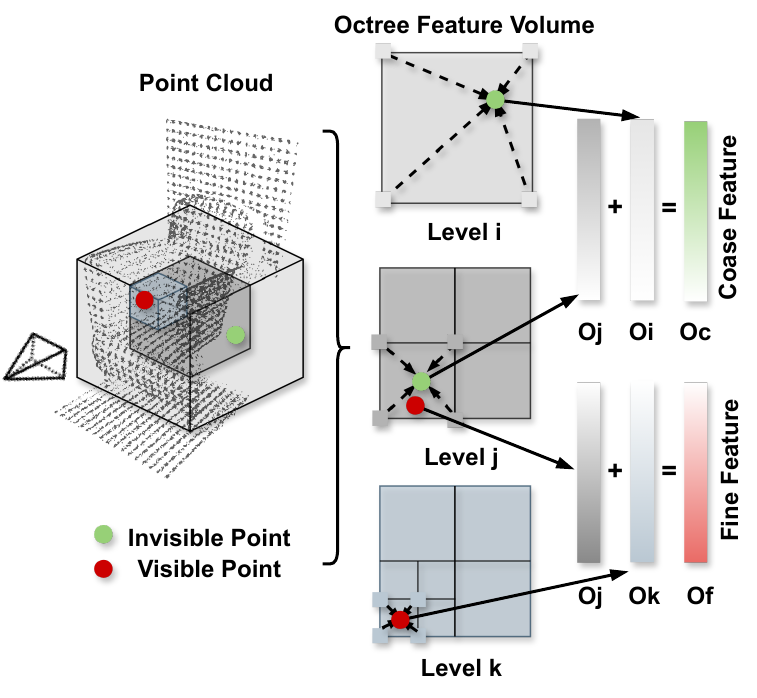}
    \vspace{-4mm}
    \caption{Coarse-fine hierarchical octree feature volume}
  \label{fig:main:octree}
  \vspace{-5mm}
\end{figure}
\begin{figure*}[!t]
    \centering
    \includegraphics[width=1.0\textwidth]{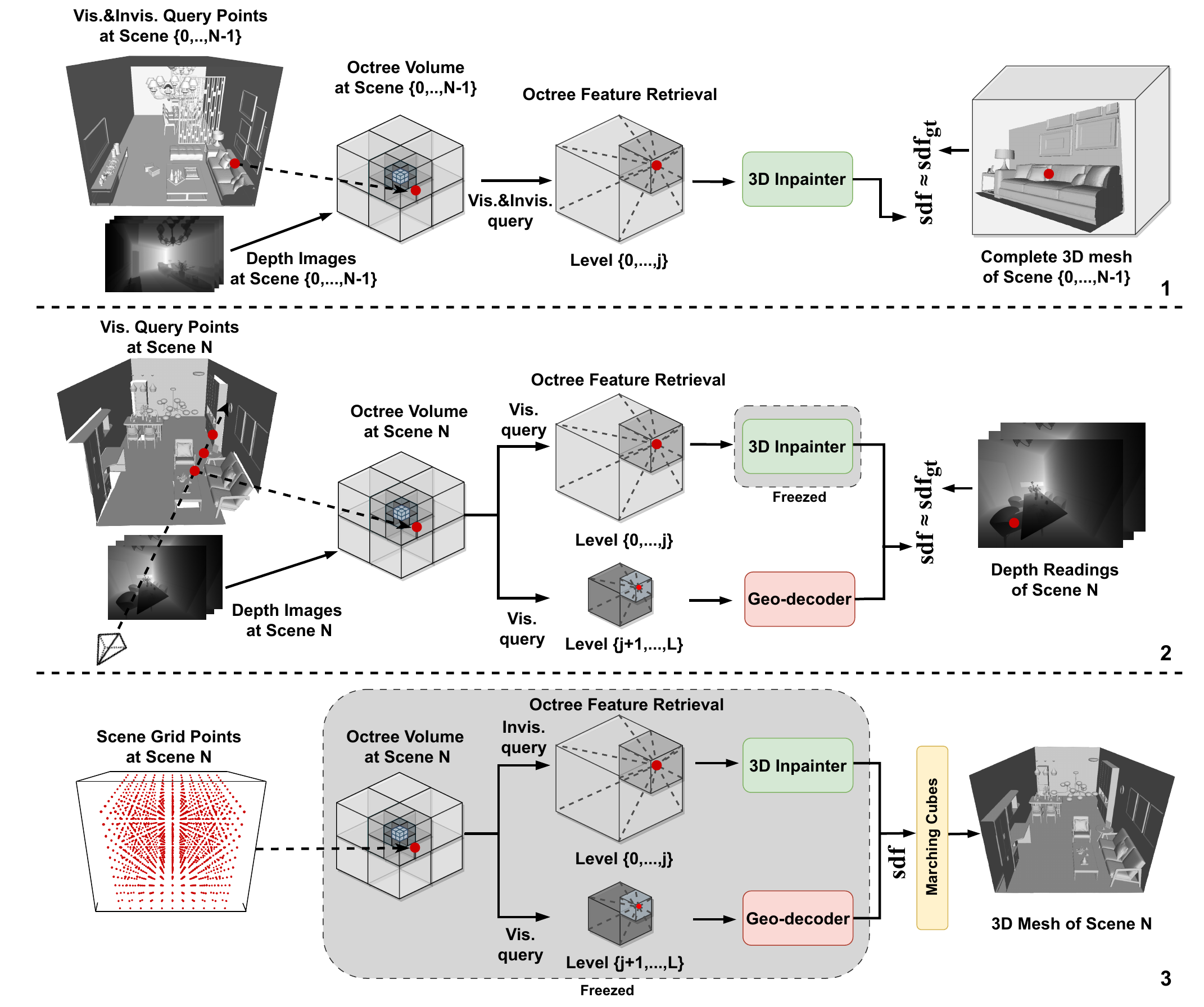}
    \vspace{-7mm}
    \caption{1: 3D Inpainter training on Scenes 0,...,N-1. The complete 3D scene meshes are provided as ground truth.  
    2: Joint Geo-decoder and octree feature volume optimization on testing Scene N. Only the visible depth images are provided as supervision, and the complete 3D scene mesh is not available. The parameters of both the Geo-decoder and octree feature volume are updated, while 3D Inpainter parameters are frozen. 
    3: 3D complete surface generation on testing Scene N.}
  \label{fig:main:pipeline}
  \vspace{-5mm}
\end{figure*}

\subsection{Overview}
Given a sequence of depth images and their associated camera poses, our method aims to reconstruct the complete 3D surfaces of a scene, including the visible surface and occluded surface.
The pipeline of our method comprising three stages is illustrated in Figure~\ref{fig:main:pipeline}. 
We first convert point clouds unprojected from given depth maps of each scene into octree-based hierarchical feature grids (defined in Sec.\ref{subsection1}). 
At the first stage, we train our 3D Inpainter in a cross-scene manner using data sampled from complete 3D meshes of scene 0,...,N-1. 
Each scene maintains its own octree feature volume while sharing the 3D Inpainter, which learns contextual structure prior (Sec.\ref{subsection2}).
At the second stage, the complete 3D mesh is not available for the unseen testing scene N and only the visible points are provided from depth images.
We jointly optimize its octree feature representation with our dual-decoder where the trained 3D Inpainter is frozen at this stage. 
The optimization relies solely on the depth observations of visible surfaces as supervision (Sec.\ref{subsection3}).
The geo-decoder with fine-level octree features is optimized using visible points. 
Meanwhile, the coarse-level octree features are optimized based on the frozen 3D inpainter, also using the available visible points.
At the third stage, using the optimized octree representation of scene N, we extract a complete 3D mesh by predicted SDFs from the dual-decoder (Sec.\ref{subsection4}).
In general, the 3D inpainter is trained using cross-scene data, enabling it to generalize on different scenes, 
while the geo-decoder is optimized using the depth observations in the test scene, which are tailored for one scene.
 
\subsection{Hierarchical Octree Feature Volume}
\label{subsection1}
Initially, we need to choose an appropriate scene representation which can encapsulate fine-grained geometric details and contextual structure information.
Similar to NGLOD~\cite{takikawa2021nglod}, we utilize learnable octree-based hierarchical feature grids to implicitly represent the entire scene's geometry, as shown in Figure~\ref{fig:main:octree}. 
%
%
Given a sequence of depth readings,  we construct a $L$ level octree, and store latent features at eight corners per octree node in each level of the tree.
We randomly initialize the corner features when creating the octree and optimize them during training. 
The octree features at level $i$ is defined as $O_{i}$. For a query point $p\in \mathbb{R}^3$ , we compute its feature vector $O_{i}(p)$ by trilinear interpolating its corresponding corner features.
%
%
%

In octree feature volume, fine-grained features that represent detailed geometric patterns are well-suited for reconstructing visible surfaces.
Coarse-level features encoding general contextual structures are more appropriate for occluded surface completion.
Therefore, we correspondingly represent visible and invisible geometry of a scene at fine and coarse levels. 
In detail, we divide the hierarchical octree features into coarse features $O_{c}$ in the high-level layer and fine feature $O_{f}$ in the low-level layer as,
$O = \{O_c, O_f\}, ~~ O_c = \{O_{i}\}_{i=0}^{i=j}, ~~ O_f = \{O_{j}\}_{i=j+1}^{i=L}$, where $j=4$ and $L=9$.
We feed the dual-decoder by concatenating features at different resolution, i.e., $O_{c}$ to 3D Inpainter and $O_{f}$ to Geo-decoder. 
The SDF values of invisible and visible surfaces are then separately inferred by the dual-decoder. 


\subsection{3D Inpainter Training Across Scenes}
\label{subsection2}
Considering invisible surface observations are not available during online optimization time, we leverage 3D complete scene meshes from multiple scenes to train a generalizable 3D Inpainter. 
We discard the widely-adopted encoder-decoder architecture used in BNV-Fusion~\cite{li2022bnv} due to its limited adaptability across different scene data. 
Instead, we adopt a decoder-only latent optimization structure, designed to learn contextual structure priors from cross-scene data.
 
Specifically, we randomly sample points $p\in \mathbb{R}^3$ including both visible $p_{v}$ and invisible points $p_{i}$, i.e., $ p=\{p_v, p_i\}$ from the complete 3D mesh of each training scene. 
We further calculate their signed distances $d_{gt}$ as ground truth based on the closest distances to the complete mesh surfaces. 
We feed both position encoding and coarse octree feature retrieved by sampled point $p$ into the 3D Inpainter $D_{Ip}$ to estimate the signed distance $d_{p}$ as, 
\vspace{-2mm}
\begin{equation}
d_{p} = D_{Ip}(f(p), O_c(p)),
\label{eq:2}
\vspace{-2mm}
\end{equation}

where $f$ refers to the position encoding function.
We train 3D Inpainter using binary cross entropy (BCE) loss $\mathcal{L}_{bce}$ as,
\vspace{-3mm}
\begin{equation}
\begin{split}
\mathcal{L}_{bce}(p) = S(d_{gt}) \cdot log(S(d_{p})) \\ 
+ (1-S(d_{gt})) \cdot log(1-S(d_{p})),
\end{split}
\label{eq:3}
\vspace{-5mm}
\end{equation}
where $S(x) = 1/(1+e^{x/\sigma})$ is sigmoid function with a flatness hyperparameter $\sigma$.
We further employ two regularization terms together with the BCE loss: eikonal term $\mathcal{L}_{eik}$~\cite{icml2020_2086} and smoothness term $\mathcal{L}_{smooth}$~\cite{Ortiz:etal:iSDF2022} as,
\vspace{-2mm}
\begin{equation}
\label{eq:4}
\mathcal{L}_{eik}(p) = (1-|| \nabla D_{Ip}(f(p), O_c(p)) ||)^2 ,
\vspace{-5mm}
\end{equation}
\vspace{-1mm}
\begin{equation}
\label{eq:5}
\begin{split}
\mathcal{L}_{smooth}(p) = || \nabla D_{Ip}(f(p), O_c(p)) \\
- \nabla D_{Ip}(f(p+\epsilon), O_c(p+\epsilon)) ||^2,
\end{split}
\vspace{-2mm}
\end{equation}
where $\epsilon$ is a small perturbation.
The global objective function $\mathcal{L}_{tr}$ of training is defined as,
\vspace{-2mm}
\begin{equation}
\label{eq:6}
\mathcal{L}_{tr}(p) = \mathcal{L}_{bce}(p) + \lambda_{eik} \mathcal{L}_{eik}(p) + \lambda_{smooth} \mathcal{L}_{smooth}(p),
\vspace{-1mm}
\end{equation}
where $\lambda_{eik}$ and $\lambda_{smooth}$ are scale factors.
During training, each scene maintains its unique octree feature volume while mutually optimizing the shared 3D Inpainter. 
We randomly select one of the scenes from training Scene 0,...,N-1 and load its feature volume. We optimize this feature volume and the 3D Inpainter for consecutive 100 iterations and repeat the process for the next selected training scene. 
The 3D Inpainter is trained to locate the high-level latent features within the octree feature volume, aggregating this contextual information to infer the SDFs of occluded regions.

\subsection{Joint Geo-decoder and Feature Optimization on New Testing Scene}
\label{subsection3} 
As the depth measurements are available in the testing scene, we can online optimize the Geo-decoder with fine-level features, which fit this specific scene for visible geometries. 
On the other hand, the coarse-level features can also be optimized using the depth reading via the frozen 3D Inpainter, which is used to infer the occluded geometries.
Note that the complete 3D mesh of the testing scene is not available at this stage.

To be specific, we sample points $p_v$ in the truncation region along the camera ray, and we further use the signed distance between the sampled point to the beam endpoint as the supervision $d_{gt}$.
During optimization, the position encoded sampled point $p_v$ is fed to both the Geo-decoder $D_{Geo}$ and 3D Inpainter $D_{Ip}$. 
The $D_{Geo}$ utilizes fine features $O_f$ retrieved by point $p_v$, while $D_{Ip}$ utilizes coarse features $O_c$ retrieved by point $p_v$ within the octree feature volume.
The estimated signed distance $d_{p_v}$ of visible point $p_v$ is obtained by
\vspace{-2mm}
\begin{equation}
d_{p_v} = D_{Geo}(f(p_v), O_f(p_v)),
\label{eq:7}
\vspace{-5mm}
\end{equation}
\begin{equation}
d_{p_v} = D_{Ip}(f(p_v), O_c(p_v)),
\label{eq:8}
\vspace{-1mm}
\end{equation}
where $f$ refers to the position encoding function. 
The global objective function $\mathcal{L}_{op}$ of optimization is defined as,
\vspace{-2mm}
\begin{equation}
\label{eq:10}
\mathcal{L}_{op}(p_v) = \mathcal{L}_{bce}(p_v) + \lambda_{eik} \mathcal{L}_{eik}(p_v) + \lambda_{smooth} \mathcal{L}_{smooth}(p_v),
\vspace{-1mm}
\end{equation}
where $\lambda_{eik}$ and $\lambda_{smooth}$ are scale factors.
Within this process, we only optimize the parameters of $D_{Geo}$, $O_f$ and $O_c$, while the parameters of $D_{Ip}$ are frozen. 
Using the scene's depth measurements, we optimize the coarse-fine feature volume tailored for the testing scene, which are used for surface generation in the next step.

\subsection{3D Surface Generation on Testing Scene}
\label{subsection4}
In the surface generation phase, we generate a complete 3D mesh using SDFs predicted from the online optimized Geo-decoder and offline trained 3D Inpainter using the optimized octree feature. 
Specifically, we uniformly sample points in the 3D space of the testing scene.
Each sampled point is used to retrieve features from all layers of the octree. 
We then apply a criterion based on the number of layers where no features are retrieved.
If the number of non-retrieved layers exceeds a threshold $\alpha=3$, the point is classified as an invisible point, denoted as $p_i$; otherwise a visible point denoted as $p_v$.
The position encoding of $p_v$ and its retrieved fine octree features $O_f$ go through the Geo-decoder to estimate the signed distance $d_{p_v}$ of the visible region as, 
\vspace{-2mm}
\begin{equation}
d_{p_v} = D_{Geo}(f(p_v), O_f(p_v)).
\label{eq:12}
\vspace{-2mm}
\end{equation}
Similar, the position encoding of $p_i$ and its retrieved coarse octree features $O_c$ go through the 3D Inpainter to estimate the signed distance $d_{p_i}$ of the occluded region as, 
\vspace{-2mm}
\begin{equation}
d_{p_i} = D_{Ip}(f(p_i), O_c(p_i)).
\label{eq:11}
\vspace{-2mm}
\end{equation}
The final SDF-assigned grid samples $d_p = \{d_{p_v}, d_{p_i}\}$ pass Marching Cube~\cite{lorensen1987marching} to produce triangle mesh. 

\section{Experiments}

\begin{table*}[!t]
\centering
\begin{tabular}{cccccc}
\hline
\textbf{Method} & \begin{tabular}[c]{@{}c@{}}Scene01\\ Accu./Comp./F1\end{tabular} & \begin{tabular}[c]{@{}c@{}}Scene05\\ Accu./Comp./F1\end{tabular} & \begin{tabular}[c]{@{}c@{}}Scene06\\ Accu./Comp./F1\end{tabular} & \begin{tabular}[c]{@{}c@{}}Scene09\\ Accu./Comp./F1\end{tabular} & \begin{tabular}[c]{@{}c@{}}Scene17\\ Accu./Comp./F1\end{tabular} \\ \hline
TSDF Fusion~\cite{Zhou2018}  & 84.3/65.8/73.9    & 92.1/66.2/77.0    & 79.2/56.0/65.6    & 81.3/51.6/63.1     &86.1/61.3/71.6                                                \\
Go-Surf~\cite{wang2022go-surf} & 88.6/75.3/81.4   &\textbf{95.8}/71.3/81.8    &\textbf{90.5}/71.7/80.0 &88.4/64.7/74.7   &87.4/71.1/78.4   \\
BNV-Fusion~\cite{li2022bnv} & \textbf{93.7}/81.7/87.3      &95.2/81.2/87.7     & 89.7/74.9/81.6 & \textbf{89.4}/68.1/77.3     &\textbf{94.4}/76.6/84.6 \\
Ours   & 91.1/\textbf{92.3}/\textbf{91.7}    & 94.3/\textbf{90.0}/\textbf{92.1}

& 88.0/\textbf{89.0}/\textbf{88.5}   & 88.2/\textbf{83.3}/\textbf{85.7}

& 90.6/\textbf{92.0}/\textbf{91.3}                                                  \\ \hline
\end{tabular}
\vspace{-3mm}
\caption{Performance comparison between the proposed method and baselines on the 3D-CRS dataset.}
\label{tab:table1}
\vspace{-2mm}
\end{table*}

\begin{table*}[!t]
\centering
\resizebox{0.999\textwidth}{!}{
\begin{tabular}{ccccccc}
\hline
\textbf{Method} & \begin{tabular}[c]{@{}c@{}}FloorPlan207\\ Accu./Comp./F1\end{tabular} & \begin{tabular}[c]{@{}c@{}}FloorPlan210\\ Accu./Comp./F1\end{tabular} & \begin{tabular}[c]{@{}c@{}}FloorPlan213\\ Accu./Comp./F1\end{tabular} & \begin{tabular}[c]{@{}c@{}}FloorPlan220\\ Accu./Comp./F1\end{tabular}& \begin{tabular}[c]{@{}c@{}}FloorPlan225\\ Accu./Comp./F1\end{tabular} & \begin{tabular}[c]{@{}c@{}}FloorPlan229\\ Accu./Comp./F1\end{tabular} \\ \hline
TSDF Fusion~\cite{Zhou2018}    & 80.3/64.2/71.3 & 86.0/68.9/76.5  & 85.8/67.9/75.8& 80.8/66.4/72.9&78.3/62.0/69.2 &76.6/67.4/71.7     \\
Go-Surf~\cite{wang2022go-surf} &  91.0/70.2/79.3  &92.2/71.0/80.2& 91.0/70.2/79.2  &83.8/67.0/74.5 & \textbf{88.8}/67.5/76.7  &86.3/73.1/79.2    						\\
BNV-Fusion~\cite{li2022bnv}      &  \textbf{92.0}/71.6/80.5& \textbf{93.3}/73.6/82.3 & \textbf{93.9}/71.9/81.4& \textbf{89.1}/70.6/78.8 &86.5/69.9/77.3  &\textbf{93.1}/76.2/83.9                                        \\
Ours            & 91.3/\textbf{90.1}/\textbf{90.7}& 92.1/\textbf{92.6}/\textbf{92.3}& 89.0/\textbf{88.3}/\textbf{88.6}& 86.0/\textbf{89.7}/\textbf{87.8}
& 87.1/\textbf{88.8}/\textbf{87.9}& 87.2/\textbf{89.4}/\textbf{88.3}                                                        \\ \hline
\end{tabular}}
\vspace{-3mm}
\caption{Performance comparison between the proposed method and baselines on the iTHOR dataset.}
\label{tab:table2}
\vspace{-4mm}
\end{table*}

\subsection{Datasets, Metrics and Baselines}
We evaluate the proposed method with the baselines on two datasets: 3D-CRS and iTHOR scene dataset from AI2-THOR~\cite{ai2thor}.
Additionally, we showcase visualization results on the real-world dataset ScanNet~\cite{dai2017scannet}.
Following BNV~\cite{li2022bnv}, the standard metrics Accuracy~(Accu.), Completeness~(Comp.) and F1 score~(F1) are employed for the quantitative analysis.
In the comparison experiments, we aimed to demonstrate the accuracy of our proposed method by comparing it against three baseline methods: TSDF-Fusion~\cite{Zhou2018}, Go-Surf~\cite{wang2022go-surf} and BNV-Fusion~\cite{li2022bnv}. 
We include more details on 3D-CRS dataset building, data preparation, metrics, baselines and network architectures in the Supplementary Material.

\subsection{Evaluation on 3D-CRS dataset}
We evaluated the proposed method by comparing it with baselines on the 3D-CRS dataset. 
We split the 20 scenes into 15 scenes for training and 5 scenes for testing.
The performance comparison with the baselines is presented in Table~\ref{tab:table1}. 
Our method outperforms the baselines significantly, particularly for the Comp. and F1 scores. 
For the Accu. score, our method achieves a competitive score with the BNV-Fusion and Go-Surf, and a much higher score than the TSDF fusion.
Comp. depicts the surface completeness of visible and occluded regions between the GT and predicted 3D mesh. 
Our method achieves a much higher Comp. score than the baselines because it reconstructs the 3D surface of occluded regions, while the baselines fail to do so. 
Accu., on the other hand, only reflects the surface accuracy of the predicted 3D surface compared with the GT 3D mesh. 
The Accu. metrics from our method encompass both visible and occluded areas, offering a more comprehensive assessment. 
In contrast, the Accu. reported by the baseline methods is limited only to the visible areas.
Our Accu. score is sometimes lower than that of the BNV-Fusion due to prediction errors in occluded regions.
The accuracy in reconstructing occluded surfaces thus contributes to a lower overall average score in our method.
Most importantly, the BNV-Fusion's encoder benefits from pre-training on the extensive ShapeNet dataset, which substantially improves its accuracy in reconstructing visible surfaces. 
In contrast, our method does not incorporate pre-training with ShapeNet data.

\begin{figure}[t]
    \centering
    \includegraphics[width=0.4\textwidth]{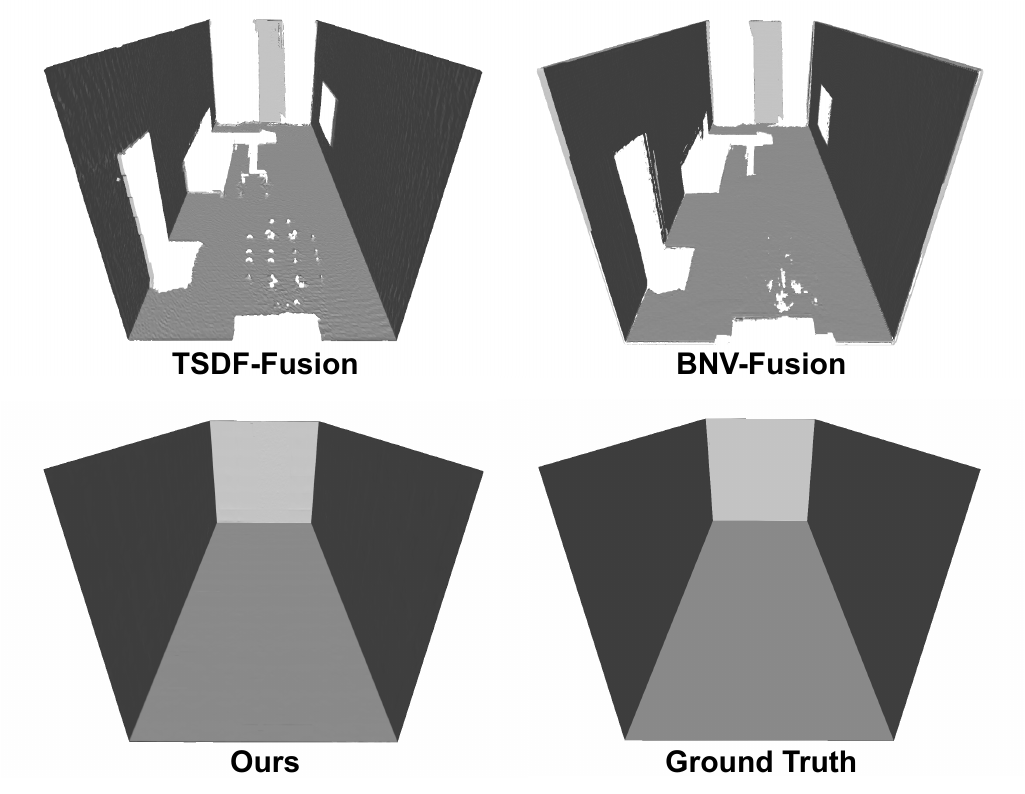}
    \vspace{-3mm}
    \caption{Visual comparison of room layout after furniture removal on the 3D-CRS dataset}
  \label{fig:main:layout}
  \vspace{-7mm}
\end{figure}

\begin{figure*}[!t]
    \centering
    \includegraphics[width=1.0\textwidth]{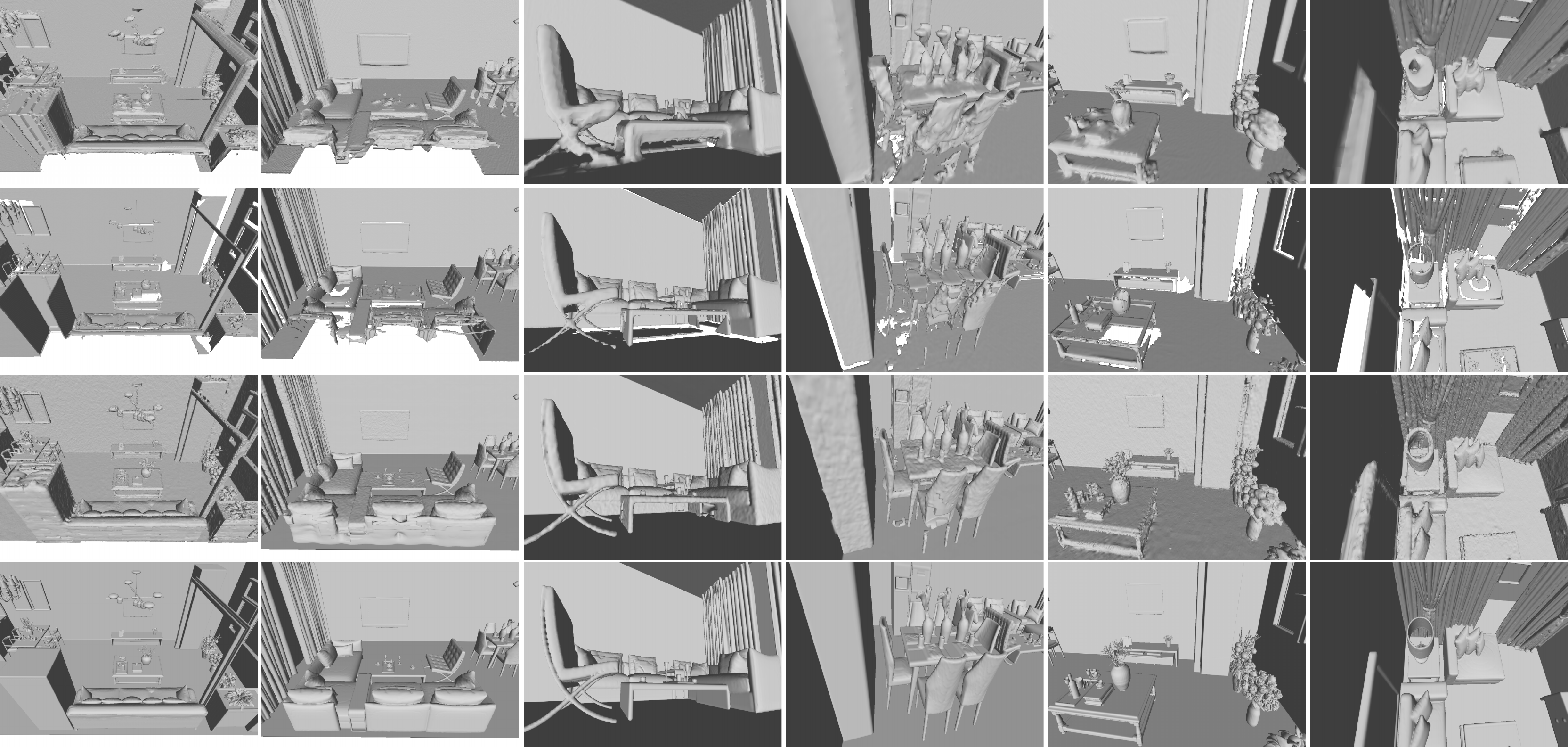}
    \vspace{-7mm}
    \caption{Visual comparison on the 3D-CRS dataset: row1-TSDF Fusion, row2-BNV Fusion, row3-Ours, row4-GT.}
  \label{fig:main:vis_CRS}
  \vspace{-3mm}
\end{figure*}

Visualizations of the 3D mesh results in Figure~\ref{fig:main:vis_CRS} reveal that the baselines could only reconstruct the surfaces of visible areas, failing to complete the surface of occluded areas. 
The TSDF-Fusion can complete smaller occluded areas but tends to inaccurately merge different surfaces, resulting in inflated surfaces. 
BNV-Fusion has a very limited ability to complete the occluded surface, especially when working with a sparse set of depth images.
In contrast, our method reconstructed the accurate and smooth 3D surface of both visible and occluded areas, particularly evident in areas like the backside of the sofa and the floor under furniture (columns 1,2,3).
Moreover, our method also provides more complete 3D shapes of complex furniture, such as tables, chairs, potted plants and ceiling lamp (columns 3,4,5,1), by accurately inpainting backside and bottom. 
These visualizations effectively highlight the superiority of our method in achieving more complete 3D surface reconstructions, where the baselines tend to leave many incomplete areas or holes.

To offer a clearer visualization of the occluded surfaces within room layouts, we present a visual comparison in Figure~\ref{fig:main:layout}, showing room layouts after furniture removal on the 3D-CRS dataset. 
It is evident that the room layout results generated by the baseline methods contain many large missing areas. 
In contrast, our method provides a complete room layout surface without obvious missing regions.
%

\subsection{Evaluation on iTHOR dataset}
\begin{figure*}[!t]
    \centering
    \includegraphics[width=1.0\textwidth]{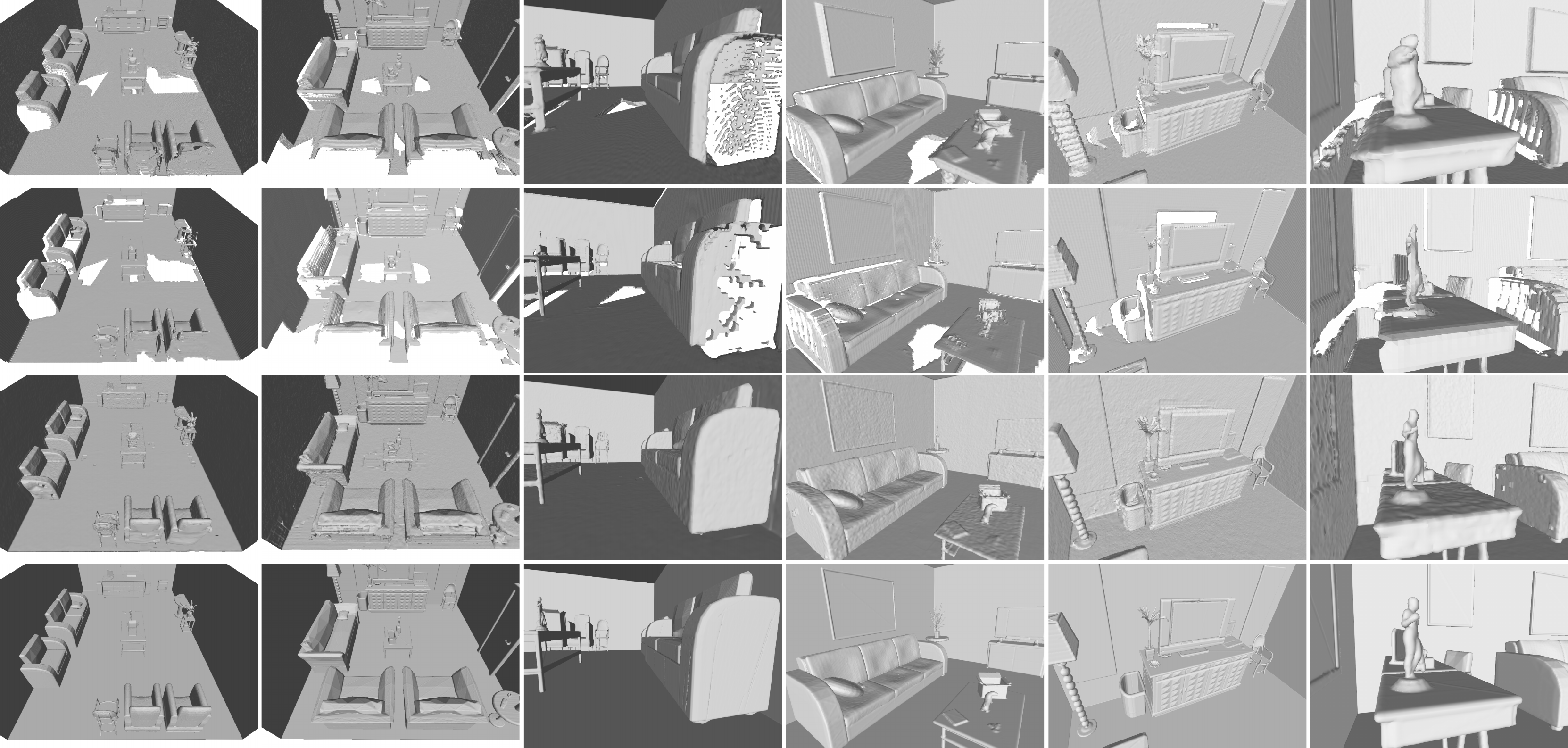}
    \vspace{-7mm}
    \caption{Visual comparison on the iTHOR dataset: row1-TSDF Fusion, row2-BNV Fusion, row3-Ours, row4-GT.}
  \label{fig:main:vis_iTHOR}
  \vspace{-5mm}
\end{figure*}

We conducted further evaluations of our proposed method against baselines on the iTHOR dataset.
We randomly selected 10 scenes from the iTHOR dataset for training and used the remaining 3 different scenes for testing.
To ensure robustness in our evaluation, we repeated this selection process twice, thereby obtaining evaluation results from a total of 6 distinct scenes.
Our evaluation includes both quantitative and qualitative comparisons, with results presented in Table~\ref{tab:table2} and Figure~\ref{fig:main:vis_iTHOR}, respectively. 
On all the testing sequences, our method outperformed the baselines on the Comp. and F1 metrics by a significant margin. 
Our method achieved a similar Accu. score to that of the BNV-Fusion and Go-Surf, while outperforming the TSDF-Fusion. 
These results demonstrate that our reconstructions are more complete, as evidenced by our high Comp. scores.

\begin{figure}[!t]
    \centering
    \includegraphics[width=0.47\textwidth]{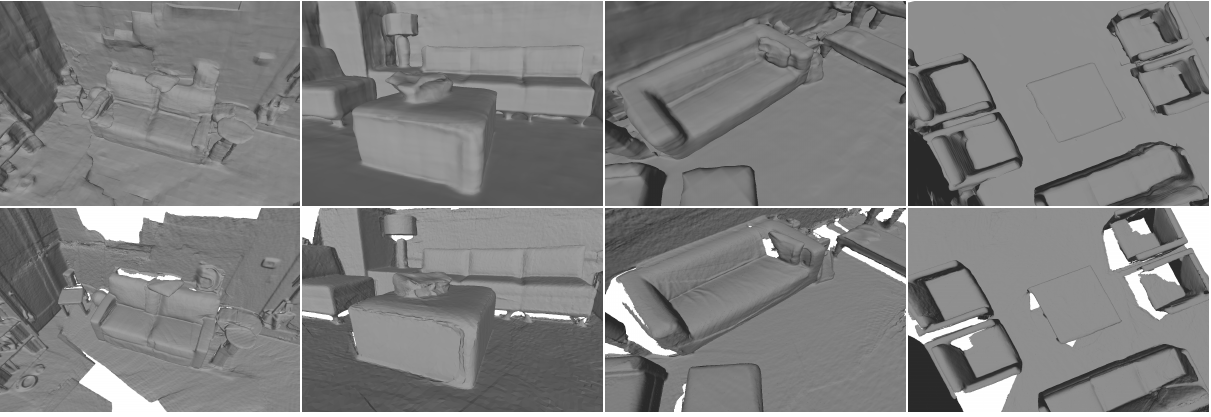}
    \vspace{-3mm}
    \caption{Visualization on ScanNet dataset: row1-Ours, row2-GT.}
    \label{fig:scannet}
    \vspace{-7mm}
\end{figure}

We present visual comparison results on the iTHOR dataset in Figure~\ref{fig:main:vis_iTHOR}. 
All the methods can reconstruct the complete 3D surface of visible areas. 
However, our method uniquely reconstructs occluded 3D surfaces, such as the unseen sides of the sofa, the wall behind the TV, and the floor beneath the furniture, which areas are missing in the baseline's reconstructions.
Most importantly, our method goes beyond just inpainting large and planar missing parts like the surfaces of the sofa, floors, and walls (column 1,2,3). 
It also accurately reconstructs high-frequency object surfaces, such as the backside of a lamp pole (column 5), the leaf of the potted plant (column 5) and the unseen parts of decorations on the table (columns 4,6).
In BNV-Fusion results, these surfaces are typically missing, while in TSDF-Fusion results, they often appear inflated.
%
%
%
%
%
%
%

\subsection{Evaluation on ScanNet dataset}
Despite the inability to conduct quantitative evaluations on the ScanNet~\cite{dai2017scannet} dataset due to their incomplete pseudo-GT 3D meshes, we qualitatively demonstrate our method's adaptability to this real-world dataset in Figure~\ref{fig:scannet}. 
Owing to ScanNet's lack of complete meshes for direct training, we trained our inpainter on 3D-CRS and tested it on ScanNet. 
The Geo-decoder and octree features were optimized using depths from the ScanNet scenes. 
Our method is observed to effectively complete areas missing from the pseudo-GT 3D mesh provided by the ScanNet.

\begin{table*}[!t]
\centering
\begin{tabular}{cccccc}
\hline
\textbf{Method} & \begin{tabular}[c]{@{}c@{}}50 Depth Frames \\ Accu./Comp./F1\end{tabular} & \begin{tabular}[c]{@{}c@{}}100 Depth Frames\\ Accu./Comp./F1\end{tabular} & \begin{tabular}[c]{@{}c@{}}150 Depth Frames\\ Accu./Comp./F1\end{tabular} & \begin{tabular}[c]{@{}c@{}}All Depth Frames\\ Accu./Comp./F1\end{tabular} \\ \hline
TSDF Fusion~\cite{Zhou2018}     	  &81.8/61.6/70.2    & 83.4/61.4/70.7	  & 84.0/61.1/70.7           &  84.6/60.2/70.2              \\
Go-Surf~\cite{wang2022go-surf}     	  &87.5/68.8/77.0    & 90.4/69.6/78.5     & 91.7/70.3/79.5           &  90.1/70.8/79.3               \\
BNV-Fusion~\cite{li2022bnv}             &\textbf{91.5}/62.3/74.0    & \textbf{91.9}/68.5/78.5    & \textbf{92.4}/74.3/82.3            &  \textbf{92.5}/76.5/83.7      		\\
Ours  &88.3/\textbf{87.5}/\textbf{87.9}   & 89.1/\textbf{88.1}/\textbf{88.6} & 89.1/\textbf{88.4}/\textbf{88.7}	 &90.4/\textbf{89.3}/\textbf{89.9}               \\ \hline
\end{tabular}
\vspace{-3mm}
\caption{Ablation study of the depth frame numbers in each scene on the 3D-CRS Dataset.}
\label{tab:table4}
\vspace{-4mm}
\end{table*}

\subsection{Ablation Study}
To demonstrate the individual effectiveness of each component in our method, we conducted ablation studies using the 3D-CRS dataset. 
Since our contributions mainly include 3D Inpainter and coarse-fine features mechanism, our ablation study analyses the impact of our designs from both aspects.
We train five different variations: 
1) our method without 3D Inpainter;
2) our method without training 3D Inpainter on cross-scene data; We optimize the 3D Inpainter together with Geo-decoder only using visible depth on testing Scene.
3) our method only uses fine features without coarse features; 
4) our method only uses the coarse features without fine features; 
5) our full method.
%
%
The average values of the evaluation metrics across five testing scenes from the 3D-CRS dataset are given in Table~\ref{tab:table3}.
In addition, we conducted an analysis to investigate how varying the number of available depth frames per scene impacts the performances of our methods in Table~\ref{tab:table4}.

\begin{table}[!t]
\resizebox{0.5\textwidth}{!}{
\begin{tabular}{clll}
\hline
\textbf{Method}                    & Accu. & Comp. & F1     \\  \hline
Ours w/o Inpainter        & \textbf{91.4}  & 82.2  & 86.5  \\
Ours w/o Cross-Scene Train Inpainter & 83.3  & 82.4  & 82.9   \\
Ours only Coarse Feature & 86.0  & 88.5  & 87.2   \\
Ours only Fine Feature   & 86.6 & 82.2 & 84.3   \\
Ours-full                & 89.1  &  \textbf{90.1}  & \textbf{89.7}  \\ \hline
\end{tabular}}
\vspace{-3mm}
\caption{Ablation study on the 3D-CRS Dataset.}
\label{tab:table3}
\vspace{-7mm}
\end{table}

\begin{figure}[!t]
\centering
\includegraphics[width=0.48\textwidth]{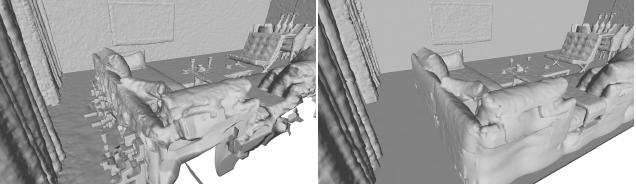}
\vspace{-7mm}
\caption{Left: 3D Inpainter w/o cross-scene training; Right: 3D Inpainter with cross-scene training.}
\label{fig:main:ablation}
\vspace{-6mm}
\end{figure}

\textbf{3D Inpainter:}
We first analyse the impact of the 3D Inpainter, specifically trained across various scenes.
From Table~\ref{tab:table3}, without 3D Inpainter or cross-scene training, these two variations achieve slightly higher Comp. scores compared to BNV-Fusion but much lower score than our full method. 
It effectively illustrates the individual contribution to the design of the 3D Inpainter. 
The first variation presents a similar Accu. score with BNV-Fusion and our full method, while the second variation exhibits a lower Accu. score. 
This is because the Accu. metric of dual-decoder reflects the average performance across both occluded and visible surface reconstructions. 
The relatively lower Accu. in reconstructing occluded surfaces thus bringing down this average value.
This also confirms that only a well-trained 3D Inpainter can accurately predict occluded surfaces.
We also provide a visual comparison in Figure~\ref{fig:main:ablation}.

\textbf{Coarse-Fine Feature:}
We secondly analyse the impact of the coarse-fine feature mechanism which distinctively represents occluded and visible regions within 3D scenes.
As shown in Table~\ref{tab:table3}, when we exclusively use either the coarse or fine features to feed into the dual-decoder, there is a noticeable decrease in both the Accu. and Comp. scores. 
When only coarse features are utilized, the performance decline can be attributed to the inadequate geometric details within coarse features, which is not precise enough for the accurate reconstruction of visible regions.
On the other hand, when solely relying on fine features, the score decrease is due to their limited coverage within the 3D space, where the 3D points sampled in the occluded regions fail to find their associated fine-level features.
Thus, the Comp. score experiences a significant drop in this case.
This limitation results in insufficient contextual structure clues within the fine features, which cannot handle the occluded surface reconstruction.
These ablation results indirectly validate the effectiveness of our coarse-fine feature mechanism, underscoring its contribution to the overall performance.

\textbf{Different Numbers of Depth Frames:}
We conducted an analysis to understand the impact of different numbers of depth frames per scene on the optimization process for the Geo-decoder and octree feature volume.
This online optimization was performed using four different sets, each comprising 50, 100, 150, and the full number of depth images, respectively.
The average values of the evaluation metrics across five testing scenes for both the baselines and our approach are given in Table~\ref{tab:table4}.
As the number of depth frames is reduced, the coverage of visible areas in the scene decreases, leading to an increase of occluded regions. 
For the baselines, their Comp. score significantly decreases due to the increase of occluded regions lacking depth measurements.
This decline occurs despite the BNV-Fusion's decoder being pre-trained on the extensive ShapeNet dataset.
Our Comp. score also remains steady with only a slight drop, even with a very limited number of depth frames. 
This consistent performance is attributed to our 3D Inpainter's capacity to accurately complete and reconstruct occluded surfaces that were invisible from the given depth frames. 
%

\section{Conclusion}
In this paper, we propose a novel indoor 3D surface reconstruction method using a sequence of depth images. 
The main new feature is that our method not only reconstructs the accurate 3D surface of the visible areas but also completes the occluded surfaces.
Our core novelty lies in the dual-decoder architecture incorporating coarse-fine hierarchical octree neural representations, designed to reconstruct occluded and visible regions respectively.
We evaluated the proposed method on two benchmarks, 3D-CRS and iTHOR, showing significant improvements in completion, especially for the occluded regions, surpassing the prior works.
The experimental results verify our method’s generalization capability of completing the 3D geometry in unseen scenes.
Lastly, we plan to release our 3D-CRS dataset for future research once its license receives approval from Bosch LLC. 

\clearpage
{
    \small
    \bibliographystyle{ieeenat_fullname}
    \bibliography{main}

\begin{thebibliography}{29}
\providecommand{\natexlab}[1]{#1}
\providecommand{\url}[1]{\texttt{#1}}
\expandafter\ifx\csname urlstyle\endcsname\relax
  \providecommand{\doi}[1]{doi: #1}\else
  \providecommand{\doi}{doi: \begingroup \urlstyle{rm}\Url}\fi

\bibitem[Azinovi{\'c} et~al.(2022)Azinovi{\'c}, Martin-Brualla, Goldman, Nie{\ss}ner, and Thies]{azinovic2022neural}
Dejan Azinovi{\'c}, Ricardo Martin-Brualla, Dan~B Goldman, Matthias Nie{\ss}ner, and Justus Thies.
\newblock Neural rgb-d surface reconstruction.
\newblock In \emph{Proceedings of the IEEE/CVF Conference on Computer Vision and Pattern Recognition}, pages 6290--6301, 2022.

\bibitem[Bokhovkin and Dai(2023)]{bokhovkin2023neural}
Aleksei Bokhovkin and Angela Dai.
\newblock Neural part priors: Learning to optimize part-based object completion in rgb-d scans.
\newblock In \emph{Proceedings of the IEEE/CVF Conference on Computer Vision and Pattern Recognition}, pages 9032--9042, 2023.

\bibitem[Curless and Levoy(1996)]{curless1996volumetric}
Brian Curless and Marc Levoy.
\newblock A volumetric method for building complex models from range images.
\newblock In \emph{Proceedings of the 23rd annual conference on Computer graphics and interactive techniques}, pages 303--312, 1996.

\bibitem[Dai et~al.(2017{\natexlab{a}})Dai, Chang, Savva, Halber, Funkhouser, and Nie{\ss}ner]{dai2017scannet}
Angela Dai, Angel~X. Chang, Manolis Savva, Maciej Halber, Thomas Funkhouser, and Matthias Nie{\ss}ner.
\newblock Scannet: Richly-annotated 3d reconstructions of indoor scenes.
\newblock In \emph{Proc. Computer Vision and Pattern Recognition (CVPR), IEEE}, 2017{\natexlab{a}}.

\bibitem[Dai et~al.(2017{\natexlab{b}})Dai, Nie{\ss}ner, Zoll{\"o}fer, Izadi, and Theobalt]{dai2017bundlefusion}
Angela Dai, Matthias Nie{\ss}ner, Michael Zoll{\"o}fer, Shahram Izadi, and Christian Theobalt.
\newblock Bundlefusion: Real-time globally consistent 3d reconstruction using on-the-fly surface re-integration.
\newblock \emph{ACM Transactions on Graphics 2017 (TOG)}, 2017{\natexlab{b}}.

\bibitem[Endres et~al.(2013)Endres, Hess, Sturm, Cremers, and Burgard]{endres20133}
Felix Endres, J{\"u}rgen Hess, J{\"u}rgen Sturm, Daniel Cremers, and Wolfram Burgard.
\newblock 3-d mapping with an rgb-d camera.
\newblock \emph{IEEE transactions on robotics}, 30\penalty0 (1):\penalty0 177--187, 2013.

\bibitem[Gropp et~al.(2020)Gropp, Yariv, Haim, Atzmon, and Lipman]{icml2020_2086}
Amos Gropp, Lior Yariv, Niv Haim, Matan Atzmon, and Yaron Lipman.
\newblock Implicit geometric regularization for learning shapes.
\newblock In \emph{Proceedings of Machine Learning and Systems 2020}, pages 3569--3579, 2020.

\bibitem[Kim et~al.(2023)Kim, Brown, Yin, Kreis, Schwarz, Li, Rombach, Torralba, and Fidler]{Kim_2023_CVPR}
Seung~Wook Kim, Bradley Brown, Kangxue Yin, Karsten Kreis, Katja Schwarz, Daiqing Li, Robin Rombach, Antonio Torralba, and Sanja Fidler.
\newblock Neuralfield-ldm: Scene generation with hierarchical latent diffusion models.
\newblock In \emph{Proceedings of the IEEE/CVF Conference on Computer Vision and Pattern Recognition (CVPR)}, 2023.

\bibitem[Kolve et~al.(2017)Kolve, Mottaghi, Han, VanderBilt, Weihs, Herrasti, Gordon, Zhu, Gupta, and Farhadi]{ai2thor}
Eric Kolve, Roozbeh Mottaghi, Winson Han, Eli VanderBilt, Luca Weihs, Alvaro Herrasti, Daniel Gordon, Yuke Zhu, Abhinav Gupta, and Ali Farhadi.
\newblock {AI2-THOR: An Interactive 3D Environment for Visual AI}.
\newblock \emph{arXiv}, 2017.

\bibitem[Li et~al.(2023)Li, Dong, Wen, Gao, Huang, Liu, and Cremers]{Li_2023_ICCV}
Haoang Li, Jinhu Dong, Binghui Wen, Ming Gao, Tianyu Huang, Yun-Hui Liu, and Daniel Cremers.
\newblock Ddit: Semantic scene completion via deformable deep implicit templates.
\newblock In \emph{Proceedings of the IEEE/CVF International Conference on Computer Vision (ICCV)}, 2023.

\bibitem[Li et~al.(2022)Li, Tang, Prisacariu, and Torr]{li2022bnv}
Kejie Li, Yansong Tang, Victor~Adrian Prisacariu, and Philip~HS Torr.
\newblock Bnv-fusion: Dense 3d reconstruction using bi-level neural volume fusion.
\newblock In \emph{Proceedings of the IEEE/CVF Conference on Computer Vision and Pattern Recognition}, pages 6166--6175, 2022.

\bibitem[Lorensen and Cline(1987)]{lorensen1987marching}
William~E Lorensen and Harvey~E Cline.
\newblock Marching cubes: A high resolution 3d surface construction algorithm.
\newblock \emph{ACM siggraph computer graphics}, 21\penalty0 (4):\penalty0 163--169, 1987.

\bibitem[Melas-Kyriazi et~al.(2023)Melas-Kyriazi, Laina, Rupprecht, and Vedaldi]{Melas-Kyriazi_2023_CVPR}
Luke Melas-Kyriazi, Iro Laina, Christian Rupprecht, and Andrea Vedaldi.
\newblock Realfusion: 360deg reconstruction of any object from a single image.
\newblock In \emph{Proceedings of the IEEE/CVF Conference on Computer Vision and Pattern Recognition (CVPR)}, pages 8446--8455, 2023.

\bibitem[Mildenhall et~al.(2021)Mildenhall, Srinivasan, Tancik, Barron, Ramamoorthi, and Ng]{mildenhall2021nerf}
Ben Mildenhall, Pratul~P Srinivasan, Matthew Tancik, Jonathan~T Barron, Ravi Ramamoorthi, and Ren Ng.
\newblock Nerf: Representing scenes as neural radiance fields for view synthesis.
\newblock \emph{Communications of the ACM}, 65\penalty0 (1):\penalty0 99--106, 2021.

\bibitem[Newcombe et~al.(2011)Newcombe, Izadi, Hilliges, Molyneaux, Kim, Davison, Kohi, Shotton, Hodges, and Fitzgibbon]{newcombe2011kinectfusion}
Richard~A Newcombe, Shahram Izadi, Otmar Hilliges, David Molyneaux, David Kim, Andrew~J Davison, Pushmeet Kohi, Jamie Shotton, Steve Hodges, and Andrew Fitzgibbon.
\newblock Kinectfusion: Real-time dense surface mapping and tracking.
\newblock In \emph{2011 10th IEEE international symposium on mixed and augmented reality}, pages 127--136. Ieee, 2011.

\bibitem[Ortiz et~al.(2022)Ortiz, Clegg, Dong, Sucar, Novotny, Zollhoefer, and Mukadam]{Ortiz:etal:iSDF2022}
Joseph Ortiz, Alexander Clegg, Jing Dong, Edgar Sucar, David Novotny, Michael Zollhoefer, and Mustafa Mukadam.
\newblock isdf: Real-time neural signed distance fields for robot perception.
\newblock In \emph{Robotics: Science and Systems}, 2022.

\bibitem[Park et~al.(2019)Park, Florence, Straub, Newcombe, and Lovegrove]{park2019deepsdf}
Jeong~Joon Park, Peter Florence, Julian Straub, Richard Newcombe, and Steven Lovegrove.
\newblock Deepsdf: Learning continuous signed distance functions for shape representation.
\newblock In \emph{Proceedings of the IEEE/CVF conference on computer vision and pattern recognition}, pages 165--174, 2019.

\bibitem[Peng et~al.(2020)Peng, Niemeyer, Mescheder, Pollefeys, and Geiger]{peng2020convolutional}
Songyou Peng, Michael Niemeyer, Lars Mescheder, Marc Pollefeys, and Andreas Geiger.
\newblock Convolutional occupancy networks.
\newblock In \emph{Computer Vision--ECCV 2020: 16th European Conference, Glasgow, UK, August 23--28, 2020, Proceedings, Part III 16}, pages 523--540. Springer, 2020.

\bibitem[Pfister et~al.(2000)Pfister, Zwicker, Van~Baar, and Gross]{pfister2000surfels}
Hanspeter Pfister, Matthias Zwicker, Jeroen Van~Baar, and Markus Gross.
\newblock Surfels: Surface elements as rendering primitives.
\newblock In \emph{Proceedings of the 27th annual conference on Computer graphics and interactive techniques}, pages 335--342, 2000.

\bibitem[Sucar et~al.(2021)Sucar, Liu, Ortiz, and Davison]{sucar2021imap}
Edgar Sucar, Shikun Liu, Joseph Ortiz, and Andrew~J Davison.
\newblock imap: Implicit mapping and positioning in real-time.
\newblock In \emph{Proceedings of the IEEE/CVF International Conference on Computer Vision}, pages 6229--6238, 2021.

\bibitem[Takikawa et~al.(2021)Takikawa, Litalien, Yin, Kreis, Loop, Nowrouzezahrai, Jacobson, McGuire, and Fidler]{takikawa2021nglod}
Towaki Takikawa, Joey Litalien, Kangxue Yin, Karsten Kreis, Charles Loop, Derek Nowrouzezahrai, Alec Jacobson, Morgan McGuire, and Sanja Fidler.
\newblock Neural geometric level of detail: Real-time rendering with implicit {3D} shapes.
\newblock \emph{Proceedings of the IEEE/CVF Conference on Computer Vision and Pattern Recognition (CVPR)}, 2021.

\bibitem[Wang et~al.(2022)Wang, Bleja, and Agapito]{wang2022go-surf}
Jingwen Wang, Tymoteusz Bleja, and Lourdes Agapito.
\newblock Go-surf: Neural feature grid optimization for fast, high-fidelity rgb-d surface reconstruction.
\newblock In \emph{2022 International Conference on 3D Vision (3DV)}. IEEE, 2022.

\bibitem[Wang et~al.(2023)Wang, Liu, Gao, Shi, Fang, and Han]{wang2023lp}
Meng Wang, Yu-Shen Liu, Yue Gao, Kanle Shi, Yi Fang, and Zhizhong Han.
\newblock Lp-dif: Learning local pattern-specific deep implicit function for 3d objects and scenes.
\newblock In \emph{Proceedings of the IEEE/CVF Conference on Computer Vision and Pattern Recognition}, pages 21856--21865, 2023.

\bibitem[Whelan et~al.(2015)Whelan, Leutenegger, Salas-Moreno, Glocker, and Davison]{whelan2015elasticfusion}
Thomas Whelan, Stefan Leutenegger, Renato Salas-Moreno, Ben Glocker, and Andrew Davison.
\newblock Elasticfusion: Dense slam without a pose graph.
\newblock In \emph{Robotics: Science and Systems}, 2015.

\bibitem[Wu et~al.(2023)Wu, Wang, Li, Zheng, and Cai]{wu2023objectsdf++}
Qianyi Wu, Kaisiyuan Wang, Kejie Li, Jianmin Zheng, and Jianfei Cai.
\newblock Objectsdf++: Improved object-compositional neural implicit surfaces.
\newblock In \emph{Proceedings of the IEEE/CVF International Conference on Computer Vision}, pages 21764--21774, 2023.

\bibitem[Ye et~al.(2023)Ye, Liu, Li, and Yang]{ye2023self}
Botao Ye, Sifei Liu, Xueting Li, and Ming-Hsuan Yang.
\newblock Self-supervised super-plane for neural 3d reconstruction.
\newblock In \emph{Proceedings of the IEEE/CVF Conference on Computer Vision and Pattern Recognition}, pages 21415--21424, 2023.

\bibitem[Yu et~al.(2022)Yu, Peng, Niemeyer, Sattler, and Geiger]{Yu2022MonoSDF}
Zehao Yu, Songyou Peng, Michael Niemeyer, Torsten Sattler, and Andreas Geiger.
\newblock Monosdf: Exploring monocular geometric cues for neural implicit surface reconstruction.
\newblock \emph{Advances in Neural Information Processing Systems (NeurIPS)}, 2022.

\bibitem[Zhou et~al.(2018)Zhou, Park, and Koltun]{Zhou2018}
Qian-Yi Zhou, Jaesik Park, and Vladlen Koltun.
\newblock {Open3D}: {A} modern library for {3D} data processing.
\newblock \emph{arXiv:1801.09847}, 2018.

\bibitem[Zhu et~al.(2022)Zhu, Peng, Larsson, Xu, Bao, Cui, Oswald, and Pollefeys]{zhu2022nice}
Zihan Zhu, Songyou Peng, Viktor Larsson, Weiwei Xu, Hujun Bao, Zhaopeng Cui, Martin~R Oswald, and Marc Pollefeys.
\newblock Nice-slam: Neural implicit scalable encoding for slam.
\newblock In \emph{Proceedings of the IEEE/CVF Conference on Computer Vision and Pattern Recognition}, pages 12786--12796, 2022.

\end{thebibliography}
}


\end{document}